\pdfoutput=1
\documentclass[10pt]{article} 

\usepackage[preprint]{rlj}

\usepackage{amssymb}            
\usepackage{mathtools}          
\usepackage{mathrsfs}           
\usepackage{graphicx}           
\usepackage{subcaption}         
\usepackage{wrapfig}            
\usepackage[space]{grffile}     
\usepackage{url}                
\usepackage{booktabs}
\usepackage{lipsum}             

\title{Better Slots, Better Worlds: Representation Quality \& Robustness in Object-Centric World Models}

\setrunningtitle{Better Slots, Better Worlds}

\author{Shukrullo Nazirjonov\textsuperscript{1}, Sai Prasanna \textsuperscript{1,2,$\dagger$}, Anna Manasyan \textsuperscript{1,2,$\dagger$}, Georg Martius \textsuperscript{1,2}}

\emails{ \ \{shukrullo.nazirjonov, sai.raman, anna.manasyan\}@uni-tuebingen.de}

\affiliations{
$^{1}$\textbf{University of Tuebingen}\\
$^{2}$\textbf{Max Planck Institute for Intelligent Systems}\\
$^\dagger$ Equal contribution
}

\contribution{
    \textbf{Quality.} Analysis of the relationship of object-centric representation quality with downstream planning performance of object-centric world models.
    }
    {
    Object-centric Encoders: Slot Contrast, Evaluation Environments: Push-T, OGBench-Cube
    }

\contribution{
    \textbf{Robustness.} Analysis of robustness in planning performance of object-centric world models under environment distribution shifts.
    }
    {
    None
    }


\summary{Learning world models from offline trajectories enables agents to accomplish different tasks through planning. Object-centric (OC) representations, which decompose a scene into a set of slots that bind to its objects, have been proposed as an inductive bias for world models that are more sample-efficient and generalize better. Yet prior object-centric world models (OCWMs) take the slot encoder as given and evaluate only in-distribution, leaving open whether the object-centric bias actually delivers for \textit{planning} and what within the OCWM drives it. We conduct a controlled study of OCWMs for visual model-predictive control along two axes: object-centric representation quality and generalization under distribution shift relative to scene-centric models. We find that (i) planning success correlates positively with unsupervised slot-quality metrics (FG-ARI, mBO), though the gains saturate at high slot quality; (ii) with well-bound slots, the auxiliary proprioception inputs and masking inductive bias that prior methods relied on become unnecessary; and (iii) an OCWM with well-bound slots plans more robustly under unseen distribution shifts than the end-to-end trained scene-centric LeWM, while DINO-WM, built on similar frozen pretrained features, remains comparably robust --- suggesting that pretrained visual representations are an important contributor to robustness.
}

\begin{document}

\maketitle  


\begin{abstract}

Learning world models from offline trajectories enables agents to accomplish different tasks through planning. Object-centric (OC) representations, which decompose a scene into a set of slots that bind to its objects, have been proposed as an inductive bias for world models that are more sample-efficient and generalize better. Yet prior object-centric world models (OCWMs) take the slot encoder as given and evaluate only in-distribution, leaving open whether the object-centric bias actually delivers for \textit{planning} and what within the OCWM drives it. We conduct a controlled study of OCWMs for visual model-predictive control along two axes: object-centric representation quality and generalization under distribution shift relative to scene-centric models. We find that (i) planning success correlates positively with unsupervised slot-quality metrics (FG-ARI, mBO), though the gains saturate at high slot quality; (ii) with well-bound slots, the auxiliary proprioception inputs and masking inductive bias that prior methods relied on become unnecessary; and (iii) an OCWM with well-bound slots plans more robustly under unseen distribution shifts than the end-to-end trained scene-centric LeWM, while DINO-WM, built on similar frozen pretrained features, remains comparably robust --- suggesting that pretrained visual representations are an important contributor to robustness.

\end{abstract}
\section{Introduction}

A central promise of world models \citep{ha2018worldmodels} is that an agent can learn its environment's dynamics from offline trajectories, and then plan toward arbitrary goals at test time \citep{zhou2024dinowmworldmodelspretrained, leworldmodel2026}. Since offline action-conditioned trajectories are costly to collect, the representation a world model uses matters for sample efficiency and generalization. Prior work spans a spectrum of representations: scene-centric models predict over DINOv2 patch features \citep{oquab2024dinov2learningrobustvisual} (DINO-WM) or a single global latent (LeWM), while object-centric models predict over a set of object slots (C-JEPA) \citep{nam2026causaljepalearningworldmodels}. Object-centric representations are appealing in principle: by factorizing a scene to match its compositional, causal structure \citep{Locatello2020SlotAttention, schoelkopf2021towards}, they should allow a world model to generalize under distribution shifts. Two assumptions behind this appeal, however, are untested. First, the slot encoder is trained separately from the world model and selected using unsupervised slot-quality metrics (FG-ARI \citep{greff2019multi} and mBO \citep{ponttuset2017multiscale}) that reward clean object masks rather than planning success; whether better-scoring slots yield better downstream planning is therefore unknown. Second, whether the promised generalization holds for planning under distribution shift is untested.

We address both with a controlled study of OCWMs for visual model-predictive control on 2D PushT and 3D OGBench-Cube, along two axes: for quality, we train world models on SlotContrast \citep{manasyan2025temporally} checkpoints of varying slot quality and measure planning success; for generalization, we compare our OCWM against scene-centric world models (DINO-WM, LeWM) under matched conditions. Our findings are:

\begin{itemize}
    \item \textbf{Quality.} Planning success correlates positively with
  unsupervised slot-quality metrics (FG-ARI, mBO), with gains saturating at high quality; with well-bound slots, the OCWM no longer needs the auxiliary proprioception inputs and masking inductive bias that prior methods relied on.
    \item \textbf{Robustness.} Under unseen distribution shifts, the OCWM degrades the least, while DINO-WM remains similarly robust and LeWM degrades substantially; together, these results suggest that planning over frozen pretrained visual representations is associated with improved robustness.
\end{itemize}
\section{Related Work}
OCWMs such as OC-STORM and C-JEPA \citep{zhang2025ocstorm,nam2026causaljepalearningworldmodels,spieler2026slot} are evaluated only in-distribution. While \cite{wang2026dyn} evaluate their Dyn-O OCWM on held-out Procgen levels, they report only rollout visual fidelity rather than control metrics (returns or planning performance). Outside of world modeling, \citet{dittadi2022generalization} provide a controlled study of OOD generalization for slot representations on downstream property prediction, and \citet{yoon2023investigation} examine pre-trained OC representations for model-free RL. Our study fills this gap for OCWMs by sweeping slot quality with a fixed dynamics model and evaluating per-object visual, frame-level, and dynamics shifts against scene-centric WMs under matched conditions.
Related works on object-centric learning and world modeling are summarized in Appendix \ref{sec:appendix:related}.

\section{Experiments and Results}
We investigate three research questions: (1)~Does slot quality impact planning success? (2)~Do better slots allow sidestepping redundant auxiliary inputs and additional training objectives? (3)~Do object-centric representations plan more robustly under distribution shifts than scene-centric ones?
\subsection{Setup}
Our setup builds on C-JEPA~\citep{nam2026causaljepalearningworldmodels}, an object-centric world model with a non-causal variant (OC-JEPA) and a causal one (C-JEPA); both encode each frame into object slots with VideoSAUR \citep{zadaianchuk2023objectcentric} and predict future slots with a transformer dynamics module. C-JEPA augments OC-JEPA with a causal inductive bias: at each step, it partitions the slots into masked and unmasked subsets and predicts the masked slots from the unmasked ones, encouraging the model to capture \textit{inter-object interactions} rather than relying on self-dynamics.

We adopt this framework for visual model-predictive control with several improvements: we replace VideoSAUR with SlotContrast \citep{manasyan2025temporally}, which provides stronger temporal consistency and eliminates the need for Hungarian matching across slots and we update DINOv2 with DINOv3 as the feature extractor. Unless noted otherwise, our default world model, \textbf{SlotContrast-WM}, is the non-causal OC-JEPA backbone with neither the slot-masking objective nor the auxiliary proprioception token. Extended details are in Appendix~\ref{sec:appendix:encoders}.

\textbf{Environments and Baselines} We evaluate our approach on the 2D PushT and 3D OGBench-Cube environments. We compare SlotContrast-WM against two scene-centric baselines: DINO-WM, which utilizes frozen DINOv2 patch tokens, and LeWM, which relies on the global CLS token of an end-to-end trained ViT \citep{dosovitskiy2021imageworth16x16words}.  Extended hyperparameters and planning configurations are in Appendix~\ref{sec:appendix:planning}.

\textbf{Metrics} We assess slot quality with two unsupervised metrics: the Foreground Adjusted Rand Index (\emph{FG-ARI}) \citep{greff2019multi}, measuring how well objects are separated into slots, and mean Best Overlap (\emph{mBO}) \citep{ponttuset2017multiscale}, an IoU-based segmentation score that assesses mask sharpness. We use their \emph{video} variants because, for world modeling, temporally consistent slot identity matters more than per-frame segmentation quality. For planning, we report task success rate, with criteria detailed in Appendix~\ref{sec:appendix:envs}.

\subsection{Does slot quality impact planning success?}

\begin{wrapfigure}[12]{r}{0.4\textwidth}
    \vspace{-\intextsep}
    \centering
    \includegraphics[width=0.38\textwidth]{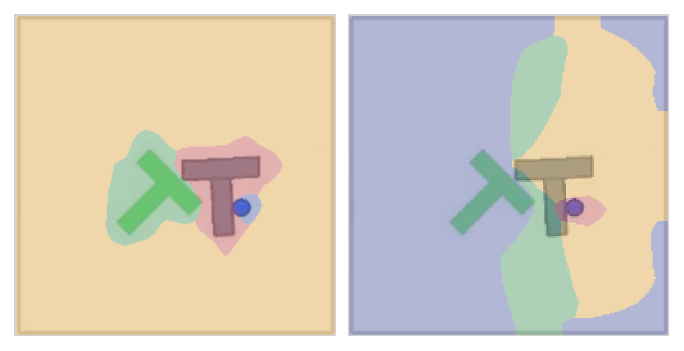}
    \caption{\small Slot decomposition on PushT. \textbf{Left:} SlotContrast binds each object (agent, T-block, goal) to a dedicated slot. \textbf{Right:} VideoSAUR fragments objects across slots and mixes them with the background.}
    \label{fig:encoder_comparison}
\end{wrapfigure}
C-JEPA, the OCWM we build on, encodes frames with a VideoSAUR encoder whose slots are poorly bound, fragmenting objects and bleeding them into the background
(Figure~\ref{fig:encoder_comparison}, right), yet C-JEPA reports strong planning performance with it. If the object-centric inductive bias is what aids planning, a representation
that so visibly fails to isolate objects ought to plan poorly, raising a more basic question:\textit{ does unsupervised slot quality bear on planning success at all}? To test this, we hold the dynamics model and planner fixed and train SlotContrast-WM on intermediate SlotContrast checkpoints spanning a range of slot quality, measuring downstream planning success.

\begin{figure}[htbp]
    \vspace{-4pt}
    \centering
    \includegraphics[width=0.7\linewidth]{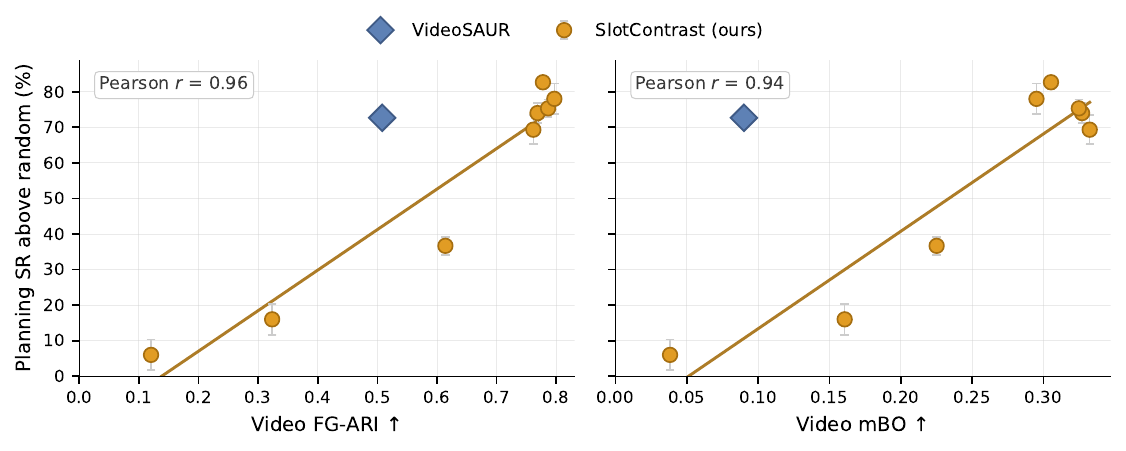}
    \caption{Across SlotContrast checkpoints (orange), planning success rate above the random-policy baseline (2\%) is positively correlated
    with video slot-quality metrics: FG-ARI (left, Pearson $r=0.96$) and mBO (right, $r=0.94$). VideoSAUR (blue diamond) uses the
    same minimal configuration (no slot masking or proprioception).}
    \label{fig:sr_vs_video_metric_best_line_pusht}
    \vspace{-12pt}
\end{figure}

As shown in Figure~\ref{fig:sr_vs_video_metric_best_line_pusht}, downstream planning performance closely tracks the emergence of object separation. On OGBench-Cube, however, the same sweep saturates early (Appendix~\ref{sec:appendix:cube_quality}, Figure~\ref{fig:sr_vs_video_cube}).

\noindent\emph{\textbf{Takeaway 1:} Planning success is positively correlated with slot quality with no change to the dynamics model, but the gains saturate: at high slot quality, where the metrics themselves plateau, better-bound slots yield little additional planning success.}

\subsection{Do better slots sidestep auxiliary inputs and training objectives?} \label{subsec:auxiliary}

Why does VideoSAUR plan well under C-JEPA despite its poorly bound slots (Figure~\ref{fig:encoder_comparison})? We hypothesize that the auxiliary mechanisms of \emph{masked-slot history prediction} objective and the \emph{auxiliary proprioception} token, which C-JEPA inherits, do not add predictive power but compensate for weak slots. We test this by sweeping \texttt{num\_masked\_slots}~$\in\{0,1,2\}$, the number of slots masked at each training step and predicted from the rest, with and without proprioception on a high-quality (SlotContrast) and a low-quality (VideoSAUR) encoder (further visualizations in Figure~\ref{fig:slot_masks_pusht}).

Without proprioception or masking, SlotContrast-WM reaches $84.7\%$ ($\pm1.9$) SR, and adding the proprioception token barely helps ($+0.6$~pp). This minimal configuration already matches, within seed variance, the $85.3\%$ ($\pm3.4$) of the full C-JEPA recipe (VideoSAUR with proprioception \emph{and} masking) and exceeds VideoSAUR ($74.7\%$) by $10$~pp. Without proprioception, masking monotonically degrades both encoders (Appendix~\ref{sec:appendix:quality}, Table~\ref{tab:masking_factorial}, $-$prop rows), and the same ordering holds on OGBench-Cube. Masking helps only when paired with proprioception on the weak encoder ($73.3\to85.3\%$), suggesting it leans on a proprioceptive shortcut rather than exploiting visual object interactions.

\noindent\emph{\textbf{Takeaway 2:} In our manipulation tasks, a sufficiently object-centric representation makes both slot history masking and proprioception unnecessary: they only compensate for weak representations. }

\subsection{Do object-centric representations allow robustness to distribution shifts?}
\begin{figure}[htbp]
    \vspace{-4pt}
    \centering
    \includegraphics[width=0.8\linewidth]{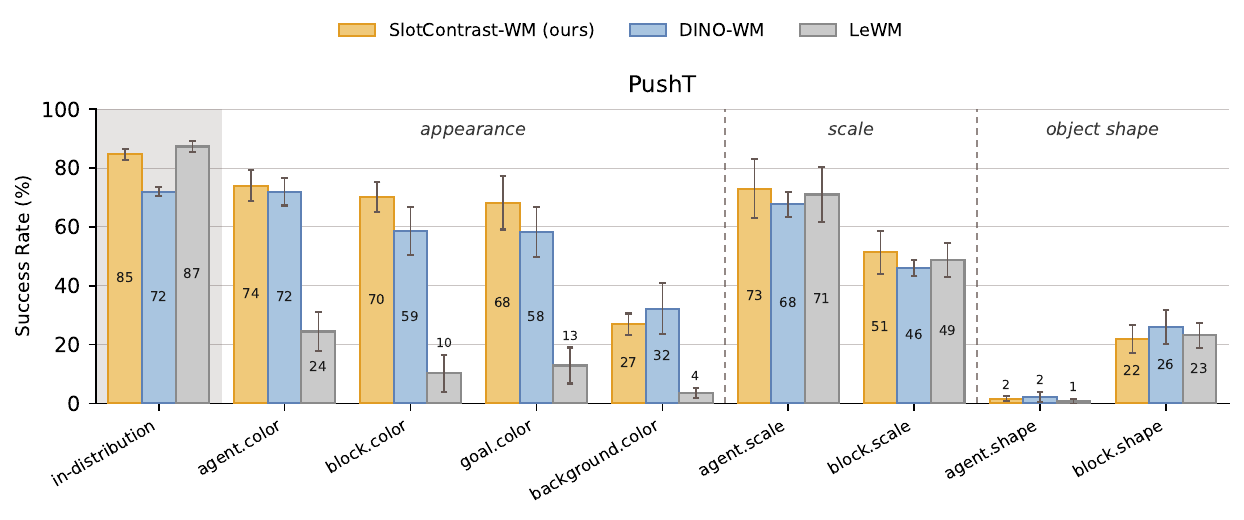}
    \caption{Planning success under distribution shift on PushT. OCWM (SlotContrast-WM) versus scene-centric baselines (DINO-WM, LeWM).}
    \label{fig:ood_generalization}
    \vspace{-12pt}
\end{figure}

We evaluate world-model robustness under visual and dynamics shifts (variations visualized in Appendix~\ref{sec:appendix:robustness}). Under object-level appearance shifts, SlotContrast-WM retains the highest success rates, and DINO-WM (patch features) degrades only moderately, whereas LeWM (CLS token of an end-to-end trained ViT) collapses (Figure~\ref{fig:ood_generalization}). DINO-WM plans over frozen pretrained DINO features directly, while SlotContrast-WM plans over object slots that Slot Attention extracts from the same features; the shared pretrained foundation appears to be the main source of robustness to appearance shifts. Frame-level perturbations such as a changed background color are difficult for all models and most damaging for LeWM, whose global CLS token they fundamentally alter. On OGBench-Cube, SlotContrast-WM and DINO-WM stay close to their in-distribution success across all shifts, while LeWM degrades under scene-level shifts (Appendix~\ref{sec:appendix:ogbench_robustness}, Figure~\ref{fig:ood_cube}). Notably, every model fails under geometric variations, since changing an object's shape alters its contact dynamics.

\noindent\emph{\textbf{Takeaway 3:} World models planning over frozen pretrained features --- object-centric or not --- tolerate appearance and frame-level shifts far better than the end-to-end trained LeWM.}

\section{Conclusion}
Our controlled study isolates what makes object-centric world models work for planning: representation quality is the dominant factor --- planning success closely tracks slot quality and saturates once slots are well-bound, at which point the auxiliary proprioception input and slot-masking objective become unnecessary.
Given good slots, the resulting OCWM is also the most robust model in our study, though DINO-WM, built on similar frozen pretrained features, retains much of this robustness while the end-to-end trained LeWM degrades severely. In short, better slots enable stronger object-centric planning, while frozen pretrained visual representations appear to be an important ingredient for robustness under distribution shift.

\textbf{Limitations \& future work.} Unsupervised slot metrics are less informative when task-relevant objects are small relative to the scene (e.g., OGBench-Cube), motivating task-aware quality measures and end-to-end encoder--world-model training. Future work should test these findings in more diverse environments—more objects, varied scales, richer dynamics—to probe OCWM's robustness and compositional generalization.



\bibliography{main}

\begin{thebibliography}{34}
\providecommand{\natexlab}[1]{#1}
\providecommand{\url}[1]{\texttt{#1}}
\expandafter\ifx\csname urlstyle\endcsname\relax
  \providecommand{\doi}[1]{DOI: #1}\else
  \providecommand{\doi}{DOI: \begingroup \urlstyle{rm}\Url}\fi

\bibitem[Aydemir et~al.(2023)Aydemir, Xie, and G{\"{u}}ney]{aydemir2023self}
G{\"{o}}rkay Aydemir, Weidi Xie, and Fatma G{\"{u}}ney.
\newblock Self-supervised object-centric learning for videos.
\newblock In Alice Oh, Tristan Naumann, Amir Globerson, Kate Saenko, Moritz
  Hardt, and Sergey Levine (eds.), \emph{Advances in Neural Information
  Processing Systems 36: Annual Conference on Neural Information Processing
  Systems 2023, NeurIPS 2023, New Orleans, LA, USA, December 10 - 16, 2023},
  2023.
\newblock URL
  \url{http://papers.nips.cc/paper\_files/paper/2023/hash/67b0e7c7c2a5780aeefe3b79caac106e-Abstract-Conference.html}.

\bibitem[Burgess et~al.(2019)Burgess, Matthey, Watters, Kabra, Higgins,
  Botvinick, and Lerchner]{burgess2019monet}
Christopher~P. Burgess, Lo{\"{\i}}c Matthey, Nicholas Watters, Rishabh Kabra,
  Irina Higgins, Matthew~M. Botvinick, and Alexander Lerchner.
\newblock Monet: Unsupervised scene decomposition and representation.
\newblock \emph{CoRR}, abs/1901.11390, 2019.
\newblock URL \url{http://arxiv.org/abs/1901.11390}.

\bibitem[Dittadi et~al.(2022)Dittadi, Papa, Vita, Sch{\"{o}}lkopf, Winther, and
  Locatello]{dittadi2022generalization}
Andrea Dittadi, Samuele~S. Papa, Michele~De Vita, Bernhard Sch{\"{o}}lkopf, Ole
  Winther, and Francesco Locatello.
\newblock Generalization and robustness implications in object-centric
  learning.
\newblock In Kamalika Chaudhuri, Stefanie Jegelka, Le~Song, Csaba
  Szepesv{\'{a}}ri, Gang Niu, and Sivan Sabato (eds.), \emph{International
  Conference on Machine Learning, {ICML} 2022, 17-23 July 2022, Baltimore,
  Maryland, {USA}}, Proceedings of Machine Learning Research, pp.\  5221--5285.
  {PMLR}, 2022.
\newblock URL \url{https://proceedings.mlr.press/v162/dittadi22a.html}.

\bibitem[Dosovitskiy et~al.(2021)Dosovitskiy, Beyer, Kolesnikov, Weissenborn,
  Zhai, Unterthiner, Dehghani, Minderer, Heigold, Gelly, Uszkoreit, and
  Houlsby]{dosovitskiy2021imageworth16x16words}
Alexey Dosovitskiy, Lucas Beyer, Alexander Kolesnikov, Dirk Weissenborn,
  Xiaohua Zhai, Thomas Unterthiner, Mostafa Dehghani, Matthias Minderer, Georg
  Heigold, Sylvain Gelly, Jakob Uszkoreit, and Neil Houlsby.
\newblock An image is worth 16x16 words: Transformers for image recognition at
  scale.
\newblock In \emph{9th International Conference on Learning Representations,
  {ICLR} 2021, Virtual Event, Austria, May 3-7, 2021}. OpenReview.net, 2021.
\newblock URL \url{https://openreview.net/forum?id=YicbFdNTTy}.

\bibitem[Elsayed et~al.(2022)Elsayed, Mahendran, van Steenkiste, Greff, Mozer,
  and Kipf]{elsayed2022savi++}
Gamaleldin~F. Elsayed, Aravindh Mahendran, Sjoerd van Steenkiste, Klaus Greff,
  Michael~C. Mozer, and Thomas Kipf.
\newblock Savi++: Towards end-to-end object-centric learning from real-world
  videos.
\newblock In Sanmi Koyejo, S.~Mohamed, A.~Agarwal, Danielle Belgrave, K.~Cho,
  and A.~Oh (eds.), \emph{Advances in Neural Information Processing Systems 35:
  Annual Conference on Neural Information Processing Systems 2022, NeurIPS
  2022, New Orleans, LA, USA, November 28 - December 9, 2022}, 2022.
\newblock URL
  \url{http://papers.nips.cc/paper\_files/paper/2022/hash/ba1a6ba05319e410f0673f8477a871e3-Abstract-Conference.html}.

\bibitem[Feng et~al.(2025)Feng, Lippe, and Magliacane]{feng2026learning}
Fan Feng, Phillip Lippe, and Sara Magliacane.
\newblock Learning interactive world model for object-centric reinforcement
  learning.
\newblock \emph{CoRR}, abs/2511.02225, 2025.
\newblock \doi{10.48550/ARXIV.2511.02225}.
\newblock URL \url{https://doi.org/10.48550/arXiv.2511.02225}.

\bibitem[Ferraro et~al.(2025)Ferraro, Mazzaglia, Verbelen, and
  Dhoedt]{ferraro2025focus}
Stefano Ferraro, Pietro Mazzaglia, Tim Verbelen, and Bart Dhoedt.
\newblock {FOCUS:} object-centric world models for robotic manipulation.
\newblock \emph{Frontiers Neurorobotics}, 19, 2025.
\newblock \doi{10.3389/FNBOT.2025.1585386}.
\newblock URL \url{https://doi.org/10.3389/fnbot.2025.1585386}.

\bibitem[Greff et~al.(2019)Greff, Kaufman, Kabra, Watters, Burgess, Zoran,
  Matthey, Botvinick, and Lerchner]{greff2019multi}
Klaus Greff, Rapha{\"{e}}l~Lopez Kaufman, Rishabh Kabra, Nick Watters, Chris
  Burgess, Daniel Zoran, Loic Matthey, Matthew~M. Botvinick, and Alexander
  Lerchner.
\newblock Multi-object representation learning with iterative variational
  inference.
\newblock In Kamalika Chaudhuri and Ruslan Salakhutdinov (eds.),
  \emph{Proceedings of the 36th International Conference on Machine Learning,
  {ICML} 2019, 9-15 June 2019, Long Beach, California, {USA}}, Proceedings of
  Machine Learning Research, pp.\  2424--2433. {PMLR}, 2019.
\newblock URL \url{http://proceedings.mlr.press/v97/greff19a.html}.

\bibitem[Ha \& Schmidhuber(2018)Ha and Schmidhuber]{ha2018worldmodels}
David Ha and J{\"{u}}rgen Schmidhuber.
\newblock Recurrent world models facilitate policy evolution.
\newblock In Samy Bengio, Hanna~M. Wallach, Hugo Larochelle, Kristen Grauman,
  Nicol{\`{o}} Cesa{-}Bianchi, and Roman Garnett (eds.), \emph{Advances in
  Neural Information Processing Systems 31: Annual Conference on Neural
  Information Processing Systems 2018, NeurIPS 2018, December 3-8, 2018,
  Montr{\'{e}}al, Canada}, pp.\  2455--2467, 2018.
\newblock URL
  \url{https://proceedings.neurips.cc/paper/2018/hash/2de5d16682c3c35007e4e92982f1a2ba-Abstract.html}.

\bibitem[Kipf et~al.(2022)Kipf, Elsayed, Mahendran, Stone, Sabour, Heigold,
  Jonschkowski, Dosovitskiy, and Greff]{kipf2022conditional}
Thomas Kipf, Gamaleldin~Fathy Elsayed, Aravindh Mahendran, Austin Stone, Sara
  Sabour, Georg Heigold, Rico Jonschkowski, Alexey Dosovitskiy, and Klaus
  Greff.
\newblock Conditional object-centric learning from video.
\newblock In \emph{The Tenth International Conference on Learning
  Representations, {ICLR} 2022, Virtual Event, April 25-29, 2022}.
  OpenReview.net, 2022.
\newblock URL \url{https://openreview.net/forum?id=aD7uesX1GF\_}.

\bibitem[Kipf et~al.(2020)Kipf, van~der Pol, and Welling]{kipf2020cswm}
Thomas~N. Kipf, Elise van~der Pol, and Max Welling.
\newblock Contrastive learning of structured world models.
\newblock In \emph{8th International Conference on Learning Representations,
  {ICLR} 2020, Addis Ababa, Ethiopia, April 26-30, 2020}. OpenReview.net, 2020.
\newblock URL \url{https://openreview.net/forum?id=H1gax6VtDB}.

\bibitem[Locatello et~al.(2020)Locatello, Weissenborn, Unterthiner, Mahendran,
  Heigold, Uszkoreit, Dosovitskiy, and Kipf]{Locatello2020SlotAttention}
Francesco Locatello, Dirk Weissenborn, Thomas Unterthiner, Aravindh Mahendran,
  Georg Heigold, Jakob Uszkoreit, Alexey Dosovitskiy, and Thomas Kipf.
\newblock Object-centric learning with slot attention.
\newblock In Hugo Larochelle, Marc'Aurelio Ranzato, Raia Hadsell,
  Maria{-}Florina Balcan, and Hsuan{-}Tien Lin (eds.), \emph{Advances in Neural
  Information Processing Systems 33: Annual Conference on Neural Information
  Processing Systems 2020, NeurIPS 2020, December 6-12, 2020, virtual}, 2020.
\newblock URL
  \url{https://proceedings.neurips.cc/paper/2020/hash/8511df98c02ab60aea1b2356c013bc0f-Abstract.html}.

\bibitem[Maes et~al.(2026{\natexlab{a}})Maes, Lidec, Haramati, Massaudi,
  Scieur, LeCun, and
  Balestriero]{maes2026stableworldmodelv1reproducibleworldmodeling}
Lucas Maes, Quentin~Le Lidec, Dan Haramati, Nassim Massaudi, Damien Scieur,
  Yann LeCun, and Randall Balestriero.
\newblock stable-worldmodel-v1: Reproducible world modeling research and
  evaluation, 2026{\natexlab{a}}.
\newblock URL \url{https://arxiv.org/abs/2602.08968}.

\bibitem[Maes et~al.(2026{\natexlab{b}})Maes, Lidec, Scieur, LeCun, and
  Balestriero]{leworldmodel2026}
Lucas Maes, Quentin~Le Lidec, Damien Scieur, Yann LeCun, and Randall
  Balestriero.
\newblock Leworldmodel: Stable end-to-end joint-embedding predictive
  architecture from pixels.
\newblock \emph{CoRR}, abs/2603.19312, 2026{\natexlab{b}}.
\newblock \doi{10.48550/ARXIV.2603.19312}.
\newblock URL \url{https://doi.org/10.48550/arXiv.2603.19312}.

\bibitem[Manasyan et~al.(2025)Manasyan, Seitzer, Radovic, Martius, and
  Zadaianchuk]{manasyan2025temporally}
Anna Manasyan, Maximilian Seitzer, Filip Radovic, Georg Martius, and Andrii
  Zadaianchuk.
\newblock Temporally consistent object-centric learning by contrasting slots.
\newblock In \emph{{IEEE/CVF} Conference on Computer Vision and Pattern
  Recognition, {CVPR} 2025, Nashville, TN, USA, June 11-15, 2025}, pp.\
  5401--5411. Computer Vision Foundation / {IEEE}, 2025.
\newblock \doi{10.1109/CVPR52734.2025.00508}.
\newblock URL
  \url{https://openaccess.thecvf.com/content/CVPR2025/html/Manasyan\_Temporally\_Consistent\_Object-Centric\_Learning\_by\_Contrasting\_Slots\_CVPR\_2025\_paper.html}.

\bibitem[Mosbach et~al.(2025)Mosbach, Ewertz, Villar{-}Corrales, and
  Behnke]{mosbach2410sold}
Malte Mosbach, Jan~Niklas Ewertz, Angel Villar{-}Corrales, and Sven Behnke.
\newblock {SOLD:} slot object-centric latent dynamics models for relational
  manipulation learning from pixels.
\newblock In Aarti Singh, Maryam Fazel, Daniel Hsu, Simon Lacoste{-}Julien,
  Felix Berkenkamp, Tegan Maharaj, Kiri Wagstaff, and Jerry Zhu (eds.),
  \emph{Forty-second International Conference on Machine Learning, {ICML} 2025,
  Vancouver, BC, Canada, July 13-19, 2025}, Proceedings of Machine Learning
  Research. {PMLR} / OpenReview.net, 2025.
\newblock URL \url{https://proceedings.mlr.press/v267/mosbach25a.html}.

\bibitem[Nam et~al.(2026)Nam, Lidec, Maes, LeCun, and
  Balestriero]{nam2026causaljepalearningworldmodels}
Heejeong Nam, Quentin~Le Lidec, Lucas Maes, Yann LeCun, and Randall
  Balestriero.
\newblock Causal-jepa: Learning world models through object-level latent
  interventions, 2026.
\newblock URL \url{https://doi.org/10.48550/arXiv.2602.11389}.

\bibitem[Oquab et~al.(2024)Oquab, Darcet, Moutakanni, Vo, Szafraniec, Khalidov,
  Fernandez, Haziza, Massa, El{-}Nouby, Assran, Ballas, Galuba, Howes, Huang,
  Li, Misra, Rabbat, Sharma, Synnaeve, Xu, J{\'{e}}gou, Mairal, Labatut,
  Joulin, and Bojanowski]{oquab2024dinov2learningrobustvisual}
Maxime Oquab, Timoth{\'{e}}e Darcet, Th{\'{e}}o Moutakanni, Huy~V. Vo, Marc
  Szafraniec, Vasil Khalidov, Pierre Fernandez, Daniel Haziza, Francisco Massa,
  Alaaeldin El{-}Nouby, Mido Assran, Nicolas Ballas, Wojciech Galuba, Russell
  Howes, Po{-}Yao Huang, Shang{-}Wen Li, Ishan Misra, Michael Rabbat, Vasu
  Sharma, Gabriel Synnaeve, Hu~Xu, Herv{\'{e}} J{\'{e}}gou, Julien Mairal,
  Patrick Labatut, Armand Joulin, and Piotr Bojanowski.
\newblock Dinov2: Learning robust visual features without supervision.
\newblock \emph{Transactions on Machine Learning Research}, 2024, 2024.
\newblock URL \url{https://openreview.net/forum?id=a68SUt6zFt}.

\bibitem[Park et~al.(2025)Park, Frans, Eysenbach, and
  Levine]{park2025ogbenchbenchmarkingofflinegoalconditioned}
Seohong Park, Kevin Frans, Benjamin Eysenbach, and Sergey Levine.
\newblock Ogbench: Benchmarking offline goal-conditioned {RL}.
\newblock In \emph{The Thirteenth International Conference on Learning
  Representations, {ICLR} 2025, Singapore, April 24-28, 2025}. OpenReview.net,
  2025.
\newblock URL \url{https://openreview.net/forum?id=M992mjgKzI}.

\bibitem[Pont{-}Tuset et~al.(2017)Pont{-}Tuset, Arbel{\'{a}}ez, Barron,
  Marqu{\'{e}}s, and Malik]{ponttuset2017multiscale}
Jordi Pont{-}Tuset, Pablo Arbel{\'{a}}ez, Jonathan~T. Barron, Ferran
  Marqu{\'{e}}s, and Jitendra Malik.
\newblock Multiscale combinatorial grouping for image segmentation and object
  proposal generation.
\newblock \emph{{IEEE} Trans. Pattern Anal. Mach. Intell.}, 39\penalty0
  (1):\penalty0 128--140, 2017.
\newblock \doi{10.1109/TPAMI.2016.2537320}.
\newblock URL \url{https://doi.org/10.1109/TPAMI.2016.2537320}.

\bibitem[Rubinstein(1999)]{Rubinstein1999}
Reuven Rubinstein.
\newblock The cross-entropy method for combinatorial and continuous
  optimization.
\newblock \emph{Methodology And Computing In Applied Probability}, 1\penalty0
  (2):\penalty0 127–190, sept 1999.
\newblock ISSN 1573-7713.
\newblock \doi{10.1023/a:1010091220143}.
\newblock URL \url{http://dx.doi.org/10.1023/A:1010091220143}.

\bibitem[Sajjadi et~al.(2022)Sajjadi, Duckworth, Mahendran, van Steenkiste,
  Pavetic, Lucic, Guibas, Greff, and
  Kipf]{sajjadi2022objectscenerepresentationtransformer}
Mehdi S.~M. Sajjadi, Daniel Duckworth, Aravindh Mahendran, Sjoerd van
  Steenkiste, Filip Pavetic, Mario Lucic, Leonidas~J. Guibas, Klaus Greff, and
  Thomas Kipf.
\newblock Object scene representation transformer, 2022.
\newblock URL
  \url{http://papers.nips.cc/paper\_files/paper/2022/hash/3dc83fcfa4d13e30070bd4b230c38cfe-Abstract-Conference.html}.

\bibitem[Sch{\"{o}}lkopf et~al.(2021)Sch{\"{o}}lkopf, Locatello, Bauer, Ke,
  Kalchbrenner, Goyal, and Bengio]{schoelkopf2021towards}
Bernhard Sch{\"{o}}lkopf, Francesco Locatello, Stefan Bauer, Nan~Rosemary Ke,
  Nal Kalchbrenner, Anirudh Goyal, and Yoshua Bengio.
\newblock Toward causal representation learning.
\newblock \emph{Proc. {IEEE}}, 109\penalty0 (5):\penalty0 612--634, 2021.
\newblock \doi{10.1109/JPROC.2021.3058954}.
\newblock URL \url{https://doi.org/10.1109/JPROC.2021.3058954}.

\bibitem[Seitzer et~al.(2023)Seitzer, Horn, Zadaianchuk, Zietlow, Xiao,
  Simon{-}Gabriel, He, Zhang, Sch{\"{o}}lkopf, Brox, and
  Locatello]{seitzer2023bridging}
Maximilian Seitzer, Max Horn, Andrii Zadaianchuk, Dominik Zietlow, Tianjun
  Xiao, Carl{-}Johann Simon{-}Gabriel, Tong He, Zheng Zhang, Bernhard
  Sch{\"{o}}lkopf, Thomas Brox, and Francesco Locatello.
\newblock Bridging the gap to real-world object-centric learning.
\newblock In \emph{The Eleventh International Conference on Learning
  Representations, {ICLR} 2023, Kigali, Rwanda, May 1-5, 2023}. OpenReview.net,
  2023.
\newblock URL \url{https://openreview.net/forum?id=b9tUk-f\_aG}.

\bibitem[Sim{\'e}oni et~al.(2025)Sim{\'e}oni, Vo, Seitzer, Baldassarre, Oquab,
  Jose, Khalidov, Szafraniec, Yi, Ramamonjisoa, Massa, Haziza, Wehrstedt, Wang,
  Darcet, Moutakanni, Sentana, Roberts, Vedaldi, Tolan, Brandt, Couprie,
  Mairal, J{\'e}gou, Labatut, and Bojanowski]{simeoni2025dinov3}
Oriane Sim{\'e}oni, Huy~V. Vo, Maximilian Seitzer, Federico Baldassarre, Maxime
  Oquab, Cijo Jose, Vasil Khalidov, Marc Szafraniec, Seungeun Yi, Micha{\"e}l
  Ramamonjisoa, Francisco Massa, Daniel Haziza, Luca Wehrstedt, Jianyuan Wang,
  Timoth{\'e}e Darcet, Th{\'e}o Moutakanni, Leonel Sentana, Claire Roberts,
  Andrea Vedaldi, Jamie Tolan, John Brandt, Camille Couprie, Julien Mairal,
  Herv{\'e} J{\'e}gou, Patrick Labatut, and Piotr Bojanowski.
\newblock Dinov3, 2025.
\newblock URL \url{https://arxiv.org/abs/2508.10104}.

\bibitem[Singh et~al.(2022)Singh, Wu, and Ahn]{Singh2022STEVE}
Gautam Singh, Yi{-}Fu Wu, and Sungjin Ahn.
\newblock Simple unsupervised object-centric learning for complex and
  naturalistic videos.
\newblock In Sanmi Koyejo, S.~Mohamed, A.~Agarwal, Danielle Belgrave, K.~Cho,
  and A.~Oh (eds.), \emph{Advances in Neural Information Processing Systems 35:
  Annual Conference on Neural Information Processing Systems 2022, NeurIPS
  2022, New Orleans, LA, USA, November 28 - December 9, 2022}, 2022.
\newblock URL
  \url{http://papers.nips.cc/paper\_files/paper/2022/hash/735c847a07bf6dd4486ca1ace242a88c-Abstract-Conference.html}.

\bibitem[Spieler et~al.(2026)Spieler, Villar-Corrales, and
  Behnke]{spieler2026slot}
Jonathan Spieler, Angel Villar-Corrales, and Sven Behnke.
\newblock Slot-mpc: Goal-conditioned model predictive control with
  object-centric representations.
\newblock \emph{arXiv preprint arXiv: 2605.14937}, 2026.

\bibitem[Veerapaneni et~al.(2019)Veerapaneni, Co{-}Reyes, Chang, Janner, Finn,
  Wu, Tenenbaum, and Levine]{veerapaneni2020op3}
Rishi Veerapaneni, John~D. Co{-}Reyes, Michael Chang, Michael Janner, Chelsea
  Finn, Jiajun Wu, Joshua~B. Tenenbaum, and Sergey Levine.
\newblock Entity abstraction in visual model-based reinforcement learning.
\newblock In Leslie~Pack Kaelbling, Danica Kragic, and Komei Sugiura (eds.),
  \emph{3rd Annual Conference on Robot Learning, CoRL 2019, Osaka, Japan,
  October 30 - November 1, 2019, Proceedings}, Proceedings of Machine Learning
  Research, pp.\  1439--1456. {PMLR}, 2019.
\newblock URL \url{http://proceedings.mlr.press/v100/veerapaneni20a.html}.

\bibitem[Wang et~al.(2026)Wang, Wang, Zhao, Stone, and Bian]{wang2026dyn}
Zizhao Wang, Kaixin Wang, Li~Zhao, Peter Stone, and Jiang Bian.
\newblock Dyn-o: Building structured world models with object-centric
  representations.
\newblock In \emph{The Thirty-ninth Annual Conference on Neural Information
  Processing Systems}, 2026.
\newblock URL \url{https://openreview.net/forum?id=b2u1yrTwFK}.

\bibitem[Wu et~al.(2023)Wu, Dvornik, Greff, Kipf, and Garg]{wu2023slotformer}
Ziyi Wu, Nikita Dvornik, Klaus Greff, Thomas Kipf, and Animesh Garg.
\newblock Slotformer: Unsupervised visual dynamics simulation with
  object-centric models.
\newblock In \emph{The Eleventh International Conference on Learning
  Representations, {ICLR} 2023, Kigali, Rwanda, May 1-5, 2023}. OpenReview.net,
  2023.
\newblock URL \url{https://openreview.net/forum?id=TFbwV6I0VLg}.

\bibitem[Yoon et~al.(2023)Yoon, Wu, Bae, and Ahn]{yoon2023investigation}
Jaesik Yoon, Yi{-}Fu Wu, Heechul Bae, and Sungjin Ahn.
\newblock An investigation into pre-training object-centric representations for
  reinforcement learning.
\newblock In Andreas Krause, Emma Brunskill, Kyunghyun Cho, Barbara Engelhardt,
  Sivan Sabato, and Jonathan Scarlett (eds.), \emph{International Conference on
  Machine Learning, {ICML} 2023, 23-29 July 2023, Honolulu, Hawaii, {USA}},
  Proceedings of Machine Learning Research, pp.\  40147--40174. {PMLR}, 2023.
\newblock URL \url{https://proceedings.mlr.press/v202/yoon23c.html}.

\bibitem[Zadaianchuk et~al.(2023)Zadaianchuk, Seitzer, and
  Martius]{zadaianchuk2023objectcentric}
Andrii Zadaianchuk, Maximilian Seitzer, and Georg Martius.
\newblock Object-centric learning for real-world videos by predicting temporal
  feature similarities.
\newblock In Alice Oh, Tristan Naumann, Amir Globerson, Kate Saenko, Moritz
  Hardt, and Sergey Levine (eds.), \emph{Advances in Neural Information
  Processing Systems 36: Annual Conference on Neural Information Processing
  Systems 2023, NeurIPS 2023, New Orleans, LA, USA, December 10 - 16, 2023},
  2023.
\newblock URL
  \url{http://papers.nips.cc/paper\_files/paper/2023/hash/c1fdec0d7ea1affa15bd09dd0fd3af05-Abstract-Conference.html}.

\bibitem[Zhang et~al.(2025)Zhang, Jelley, McInroe, and
  Storkey]{zhang2025ocstorm}
Weipu Zhang, Adam Jelley, Trevor McInroe, and Amos~J. Storkey.
\newblock Objects matter: object-centric world models improve reinforcement
  learning in visually complex environments, 2025.
\newblock URL \url{https://doi.org/10.48550/arXiv.2501.16443}.

\bibitem[Zhou et~al.(2025)Zhou, Pan, LeCun, and
  Pinto]{zhou2024dinowmworldmodelspretrained}
Gaoyue Zhou, Hengkai Pan, Yann LeCun, and Lerrel Pinto.
\newblock {DINO-WM:} world models on pre-trained visual features enable
  zero-shot planning, 2025.
\newblock URL \url{https://proceedings.mlr.press/v267/zhou25t.html}.

\end{thebibliography}
\bibliographystyle{rlj}

\newpage
\appendix

\section{Experiment Details}
\label{sec:appendix:experiments}

\subsection{Environments and Datasets}
\label{sec:appendix:envs}

We evaluate our framework across two distinct physical manipulation domains. 

\paragraph{PushT} 
A 2D planar manipulation task where a circular agent must push a T-shaped block to align with a specific target pose. 

\begin{itemize}
    \item \textbf{Dataset:} The offline dataset consists of 18,500 expert demonstrations augmented with varying levels of action noise. This dataset is adopted from \citet{leworldmodel2026}, utilizing the original generation protocol introduced by \citet{zhou2024dinowmworldmodelspretrained}.
    \item \textbf{Success Criterion:} For goal-conditioned planning, an episode is marked successful if the combined spatial translation error of both the agent and the block is less than 20 pixels, and the block's angular orientation error is less than $\pi/9$ radians ($\approx 20^\circ$) relative to the goal configuration. This follows the evaluation protocols established in prior work \citep{zhou2024dinowmworldmodelspretrained, leworldmodel2026}.
\end{itemize}

\paragraph{OGBench-Cube} 
A 3D object manipulation environment introduced by \citet{park2025ogbenchbenchmarkingofflinegoalconditioned}. The workspace features a simulated UR5e robotic arm equipped with a Robotiq 2F-85 adaptive gripper, tasked with interacting with a colored cube to achieve a spatial goal configuration.

\begin{itemize}
    \item \textbf{Dataset:} The offline dataset contains 10,000 heuristic-generated demonstrations (200 transitions per episode), adopted from \citet{leworldmodel2026}. 
    \item \textbf{Success Criterion:} An episode is considered successful if the Euclidean distance between the cube's final position and its goal target position is less than 4 cm. In contrast to the PushT, this environment does not enforce kinematic constraints on the final pose of the agent (robotic end-effector), making the random policy achieve high planning success rate. This setup follows the baseline methods and is a known limitation of the environment evaluation.
\end{itemize}
Both environments are implemented in \texttt{stable-worldmodel} \citep{maes2026stableworldmodelv1reproducibleworldmodeling} and we utilize this framework for all evaluations throughout this paper.

\subsection{Object-Centric Encoders}
\label{sec:appendix:encoders}

\paragraph{SlotContrast.}
SlotContrast \citep{manasyan2025temporally} is a video object-centric learning
framework designed to discover temporally consistent object representations from
videos. The model comprises three main components: (1) a pre-trained
self-supervised dense feature encoder, such as DINOv2
\citep{oquab2024dinov2learningrobustvisual}, which extracts rich spatial features
from each frame; (2) a Recurrent Slot Attention
\citep{Locatello2020SlotAttention, kipf2022conditional} module that groups these
features into object-centric slots while modeling their temporal evolution across
frames; and (3) a decoder that reconstructs the dense self-supervised features
from the slot representations. Training is guided by two complementary objectives.
A feature reconstruction loss encourages the slots to capture the information
necessary to reconstruct the encoder features, while a temporal consistency loss
contrasts slot representations across consecutive frames. This temporal
contrastive objective promotes stable slot assignments and enables the model to
learn object representations that remain consistent over time, leading to improved
object discovery and tracking in dynamic video scenes.

\paragraph{Our setup.}
We adopt SlotContrast as our object-centric backbone, replacing the original
DINOv2 encoder with the more recent DINOv3 \citep{simeoni2025dinov3}, which
provides stronger dense features. We otherwise follow the original training
objectives. The full list of hyperparameters for PushT and OGBench-Cube is
given in Table~\ref{tab:hyperparams}.

\begin{table}[htbp]
\centering
\caption{Hyperparameters of SlotContrast Model on PushT and OGBench-Cube.}
\label{tab:hyperparams}
\begin{tabular}{lcc}
\toprule
\textbf{Hyperparameter} & \textbf{PushT} & \textbf{OGBench} \\
\midrule
Training Steps & 100k & 100k \\
Batch Size & 128 & 128 \\
Training Segment Length & 4 & 4 \\
Learning Rate Warmup Steps & 2500 & 2500 \\
Optimizer & Adam & Adam \\
Peak Learning Rate & 0.0004 & 0.0004 \\
ViT Architecture & DINOv3 Small & DINOv3 Small \\
Initialization & FixedLearnedInit & FixedLearnedInit \\
Patch Size & 16 & 16 \\
Feature Dimension ($D_\mathrm{feat}$) & 384 & 384 \\
Gradient Norm Clipping & 0.05 & 0.05 \\
\midrule
\multicolumn{3}{l}{\textbf{Image Specifications}} \\
Image / Crop Size & 256 & 256 \\
Cropping Strategy & Full & Full \\
Augmentations & Rand. Horizontal Flip & Rand. Horizontal Flip \\
Image Tokens & 256 & 256 \\
\midrule
\multicolumn{3}{l}{\textbf{Slot Attention}} \\
Slots & 4 & 3 \\
Iterations (first / other frames) & 3 / 2 & 3 / 2 \\
Slot Dimension ($D_\mathrm{slots}$) & 128 & 128 \\
\midrule
\multicolumn{3}{l}{\textbf{Predictor}} \\
Type & Transformer & Transformer \\
Layers & 1 & 1 \\
Heads & 4 & 4 \\
\midrule
\multicolumn{3}{l}{\textbf{Decoder}} \\
Type & MLP & MLP \\
\midrule
\multicolumn{3}{l}{\textbf{Loss Parameters}} \\
Softmax Temperature ($\tau$) & 0.1 & 0.01 \\
Slot--Slot Contrast Weight ($\alpha$) & 0.01 & 0.001 \\
\bottomrule
\end{tabular}
\end{table}

\begin{figure}[htbp]

    \centering
    \begin{subfigure}[b]{\linewidth}
        \centering
        \includegraphics[width=\linewidth]{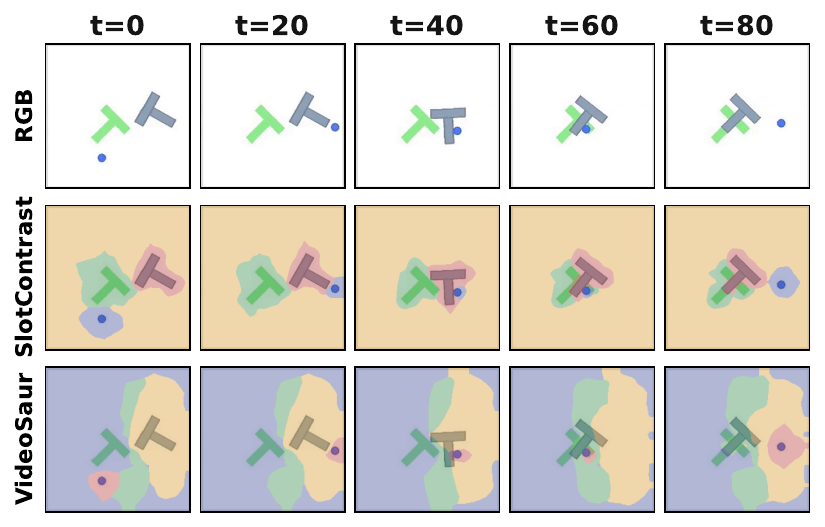}
        \caption{PushT Slot Masks: SlotContrast accurately isolates the physical objects (agent, T-block) from the background. VideoSAUR produces entangled partitions where object features bleed across slots.}
        \label{fig:slot_masks_pusht}
    \end{subfigure}
    
    \vspace{0.6cm}
    
    \begin{subfigure}[b]{\linewidth}
        \centering
        \includegraphics[width=\linewidth]{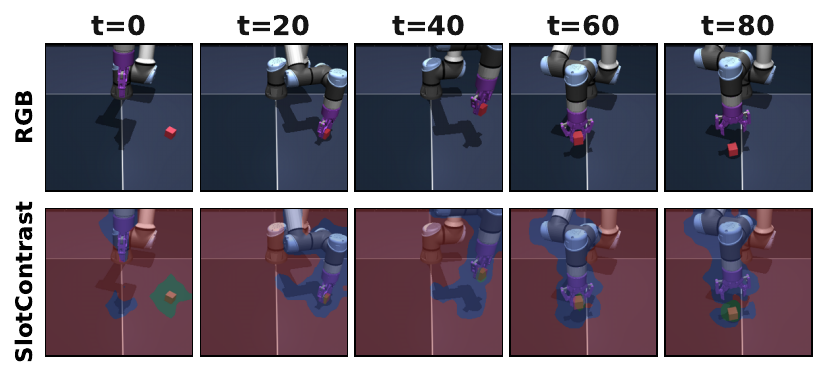}
        \caption{OGBench-Cube Slot Masks: SlotContrast successfully discovers and tracks 3D components (robotic arm and its shadow, cube, background) with temporal consistency over the episode. Notably, robotic arm and its shadow are assigned to the same slot. }
        \label{fig:slot_masks_ogbench}
    \end{subfigure}
    \caption{Qualitative comparison of learned object-centric representations over $80$ timesteps.}
    \label{fig:slot_masks}
\end{figure}

\paragraph{VideoSAUR.}
VideoSAUR \citep{zadaianchuk2023objectcentric} is another object-centric video
representation learning framework that combines DINO-based semantic features with
Slot Attention for object discovery. Its key contribution is a temporal feature
similarity objective that exploits motion bias to promote object disentanglement
by preserving semantic and temporal relationships among DINO patch embeddings.
This objective can be used either alongside a reconstruction loss or as a
standalone training signal. However, because VideoSAUR does not explicitly
constrain slot identities to remain temporally consistent, slot correspondences
across frames may need to be recovered post hoc using Hungarian matching or
similar methods.
\paragraph{Qualitative Mask Analysis.}
To visually validate the differences in slot quality between the evaluated encoders, we extract and render the learned slot masks over an episode. As shown in Figure~\ref{fig:slot_masks_pusht}, SlotContrast successfully isolates the agent, the T-block, and the background into distinct, temporally consistent slots in the PushT environment. In contrast, VideoSAUR suffers from severe feature entanglement, partitioning the space into Voronoi-like regions where object identities bleed across slot boundaries (a known limitation of Slot Attention-based methods \citep{sajjadi2022objectscenerepresentationtransformer}). Figure~\ref{fig:slot_masks_ogbench} further validates SlotContrast's object-centricity in 3D settings, demonstrating clean isolation and tracking of the robotic arm, the   gripper, and the cube, even in contact events when the end-effector grasps the cube.

\subsection{Visual Model-Based Planning}
\label{sec:appendix:planning}
All world models are evaluated under the same Cross-Entropy Method (CEM) \citep{Rubinstein1999} planner budget and identical context history.

\paragraph{Planner Cost Functions}
The CEM planner optimizes action sequences by minimizing the distance between the model's predicted latent state $\hat{Z}$ and the target goal state $Z_\text{goal}$. While all models utilize an $L_2$ norm formulation, the structure of the representation dictates exactly how this cost is computed:

\begin{itemize}
    \item \textbf{LeWM:} Evaluates the distance between the single, global \texttt{[CLS]} vectors: 
    $$J = ||\hat{z}^\text{cls} - z^\text{cls}_\text{goal}||_2^2$$
    
    \item \textbf{DINO-WM:} Evaluates the mean squared error across the rigid spatial grid of $P$ patches: 
    $$J = \frac{1}{P} \sum_{p=1}^P ||\hat{z}_p - z_{p,\text{goal}}||_2^2$$
    
    \item \textbf{SlotContrast-WM (Ours):} Because SlotContrast enforces strict temporal consistency, slot identities are preserved across time. This allows us to compute a direct, one-to-one distance across the $K$ slots without any heuristic matching: 
    $$J = \frac{1}{K} \sum_{k=1}^K ||\hat{z}_k - z_{k,\text{goal}}||_2^2$$
    
    \item \textbf{C-JEPA (VideoSAUR):} VideoSAUR does not guarantee temporal consistency. Therefore, evaluating the cost requires solving a bipartite matching problem via the Hungarian algorithm at every single planning step to align predicted slots to goal slots, where $\Pi$ is the set of all possible permutations:
    $$J = \min_{\pi \in \Pi} \frac{1}{K} \sum_{k=1}^K ||\hat{z}_k - z_{\pi(k),\text{goal}}||_2^2$$
\end{itemize}

Where specified, this visual cost is linearly combined with an auxiliary proprioceptive cost, scaled by \texttt{proprio\_cost\_weight} ($W \in \{0, 1\}$), detailed further in Appendix~\ref{sec:appendix:masking}. 

All reported planning results are averaged over three random seeds. The fixed hyperparameters used for the CEM optimization per environment are detailed in Table~\ref{tab:cem_hyperparameters}.
\begin{table}[htbp]
    \caption{Cross-Entropy Method (CEM) planning parameters used across all evaluated world models.}
    \label{tab:cem_hyperparameters}
    \begin{center}
        \begin{tabular}{lcc}
            \multicolumn{1}{l}{\bf Parameter} & \multicolumn{1}{c}{\bf PushT} & \multicolumn{1}{c}{\bf OGBench-Cube} \\ 
            \hline \\
            Environment Steps Horizon (H) & $25$ & $25$ \\
            Number of evaluated goals & $50$ & $50$ \\
            Frameskip & $5$ & $5$ \\
            Latent Planning Horizon ($H$, frameskipped) & $5$ & $5$ \\
            CEM Samples ($N_\text{samples}$) & $300$ & $300$ \\
            CEM Iterations ($N_\text{iter}$) & $30$ & $30$ \\
            Top Elites ($K$) & $30$ & $30$ \\
            Initial Sampling Variance & $1.0$ & $1.0$ \\
            MPC Execution Steps (frameskipped) & $5$ & $5$ \\
        \end{tabular}
    \end{center}
\end{table}
\section{Analyses on Encoder Quality}
\label{sec:appendix:quality}

 \subsection{Slot Quality vs.\ Planning on OGBench-Cube}
\label{sec:appendix:cube_quality}
\begin{figure}[htbp]
  \centering
\includegraphics[width=0.8\linewidth]{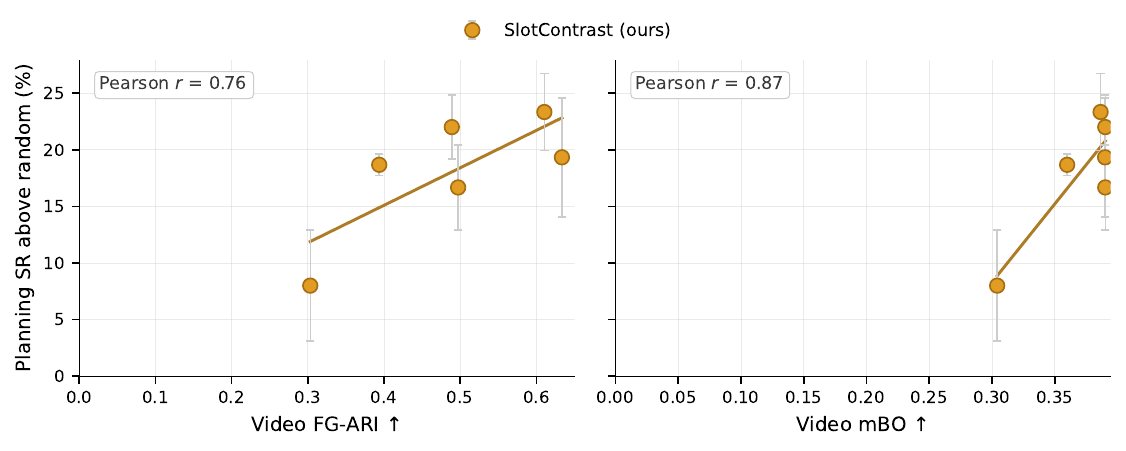}
\caption{Slot quality vs.\ planning success on OGBench-Cube. Success rate is reported relative to the random-policy baseline (48\%)}
  \label{fig:sr_vs_video_cube}
\end{figure}

Figure~\ref{fig:sr_vs_video_cube} repeats the slot-quality sweep of the main text
(Figure~\ref{fig:sr_vs_video_metric_best_line_pusht}) on OGBench-Cube. Unlike PushT, the metrics start high and the relationship
saturates early: FG-ARI and mBO are aggregated over the whole scene, where the large, easily segmented robot arm dominates the
score while the small, task-relevant cube contributes little. 

\subsection{Role of Slot Masking and Proprioception Token }
\label{sec:appendix:masking}

We sweep \texttt{num\_masked\_slots}~$\in\{0,1,2\}$ crossed with \texttt{use\_proprio}~$\in\{$true, false$\}$, separately for the two encoder families. Each cell trains a C-JEPA WM with respective object-centric encoder(SlotContrast, VideoSAUR) on the same expert dataset with the same hyperparameters.

\begin{table}[htbp]
    \caption{Planning SR (\%) on PushT evaluating \texttt{num\_masked\_slots} and proprioception across strong (SlotContrast) and weak (VideoSAUR) encoders. Without proprioception ($-$prop), masking strictly degrades planning for both encoders.}
    \label{tab:masking_factorial}
    \begin{center}
    \small
    \begin{tabular}{llccc}
        \multicolumn{1}{l}{\bf Encoder} & \multicolumn{1}{l}{\bf Proprio} & \multicolumn{1}{c}{\bf nms=0} & \multicolumn{1}{c}{\bf nms=1} & \multicolumn{1}{c}{\bf nms=2}
        \\ \hline \\
        SlotContrast & $+$prop & \textbf{85.3}\,{\scriptsize$\pm$3.4} & 82.7\,{\scriptsize$\pm$1.9} & 82.7\,{\scriptsize$\pm$3.4} \\
        SlotContrast & $-$prop & \textbf{84.7}\,{\scriptsize$\pm$1.9} & 72.0\,{\scriptsize$\pm$3.3} & 34.0\,{\scriptsize$\pm$6.5} \\
        VideoSAUR    & $+$prop & 73.3\,{\scriptsize$\pm$5.7} & \textbf{85.3}\,{\scriptsize$\pm$3.4} & 83.3\,{\scriptsize$\pm$4.1} \\
        VideoSAUR    & $-$prop & \textbf{74.7}\,{\scriptsize$\pm$1.9} & 52.0\,{\scriptsize$\pm$5.9} & 49.3\,{\scriptsize$\pm$6.6} \\
    \end{tabular}
    \end{center}
\end{table}

The same ordering holds on OGBench-Cube. With the SlotContrast-WM and no proprioception, masking provides no clear benefit in planning performance: nms=0 reaches $72.7 \pm 3.4\%$ SR versus $70.0 \pm 2.8\%$ for nms=1. 




\section{More Visualizations and Robustness Details}
\label{sec:appendix:robustness}

\subsection{OOD Variation Groups}
\label{sec:appendix:robustness:taxonomy}

Each world model is trained on a single in-distribution rendering and evaluated on a fixed grid of unseen variations, changing one factor at a time. We use the same grouping as in the main-text figures. PushT variations are grouped into \emph{appearance} (object and background colour), \emph{scale} (object size), and \emph{object shape}. OGBench-Cube variations are grouped into \emph{object-level} (cube and agent) and \emph{scene-level} (floor, camera, and lighting). The exact factors and values are listed in Tables~\ref{tab:variation_values_pusht} and~\ref{tab:variation_values_cube}.

\subsection{PushT Variation Suite}
\label{sec:appendix:robustness:pusht}

\begin{figure}[htbp]
    \centering
    \includegraphics[width=\linewidth]{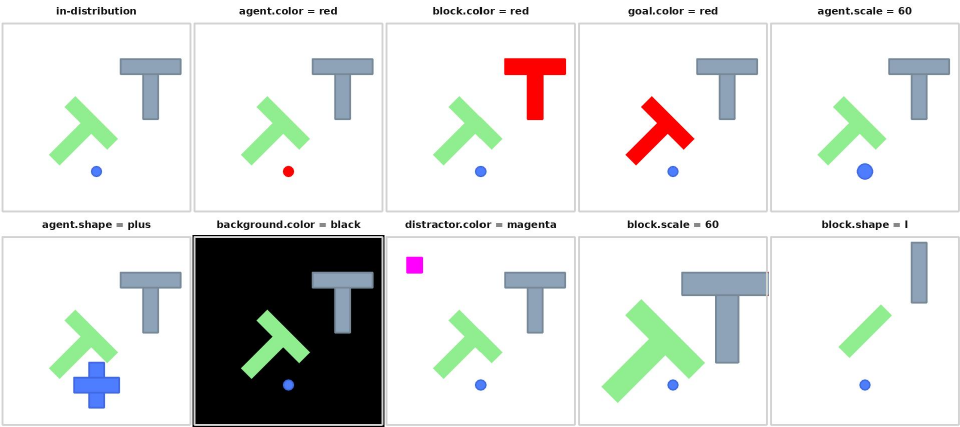}
    \caption{PushT OOD variation visualizations baseline plus one representative variation per factor. All tiles share the same fixed initial layout. Full value list is available in Table~\ref{tab:variation_values_pusht}.}
    \label{fig:variation_palette_pusht}
\end{figure}

Figure~\ref{fig:variation_palette_pusht} shows one representative variation per factor; Table~\ref{tab:variation_values_pusht} lists all values.

\begin{table}[h]
    \caption{PushT OOD variation values. ``In-dist.''\ is the training-distribution default.}
    \label{tab:variation_values_pusht}
    \begin{center}
    \small
    \begin{tabular}{llll}
        \multicolumn{1}{l}{\bf Group} & \multicolumn{1}{l}{\bf Factor} & \multicolumn{1}{l}{\bf In-dist.\ value} & \multicolumn{1}{l}{\bf OOD values}
        \\ \hline \\
        appearance   & \texttt{agent.color}      & blue        & red, yellow, black \\
                     & \texttt{block.color}      & slate gray  & red, blue, black \\
                     & \texttt{goal.color}       & light green & red, blue, black \\
                     & \texttt{background.color} & white       & red, blue, black \\
                     & \texttt{distractor.color} & off         & gray, magenta \\
        \\ \hline \\
        scale        & \texttt{agent.scale}      & $40$        & $20$, $60$ \\
                     & \texttt{block.scale}      & $40$        & $20$, $60$ \\
        \\ \hline \\
        object shape & \texttt{agent.shape}      & circle      & L, square, plus \\
                     & \texttt{block.shape}      & T           & L, square, I \\
    \end{tabular}
    \end{center}
\end{table}
\subsection{OGBench-Cube Variation Suite}
\label{sec:appendix:robustness:cube}

\begin{figure}[h]
  \centering
  \includegraphics[width=\linewidth]{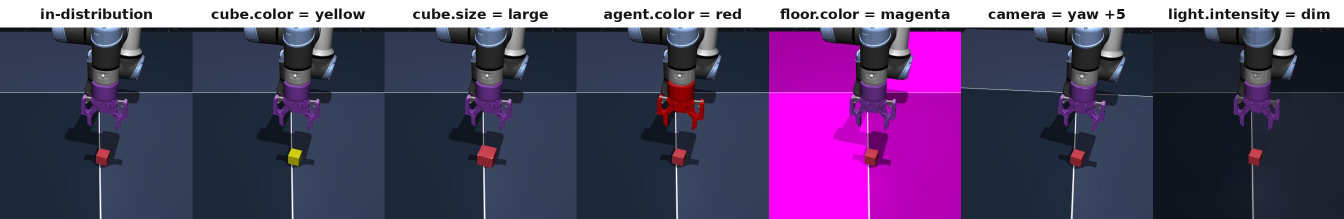}
  \caption{OGBench-Cube OOD variation visualizations: baseline and one representative variation per factor. Full
value list is available in Table~\ref{tab:variation_values_cube}.}
  \label{fig:variation_palette_cube}
\end{figure}

Figure~\ref{fig:variation_palette_cube} shows one representative variation per factor; Table~\ref{tab:variation_values_cube} lists all values.
 
\subsection{OGBench-Cube Robustness Results}
\label{sec:appendix:ogbench_robustness}

\begin{figure}[htbp]
    \centering
    \includegraphics[width=0.9 \linewidth]{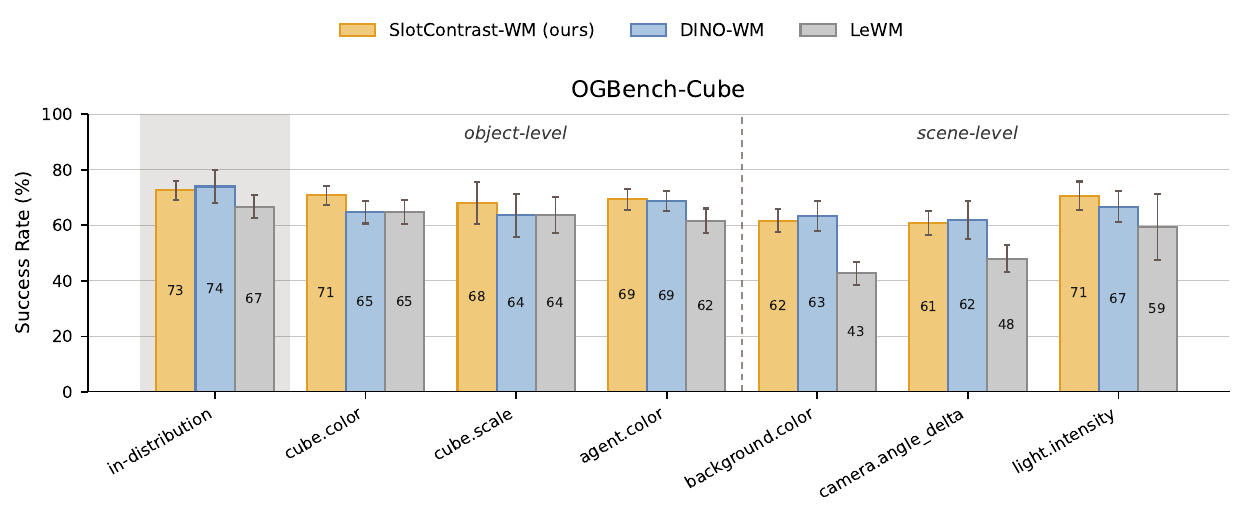}
    \caption{Zero-shot planning success rate (\%) on OGBench-Cube under object-level and scene-level distribution shifts, for SlotContrast-WM, DINO-WM, and LeWM. The random-policy baseline in this environment achieves $48\%$.}
    \label{fig:ood_cube}
\end{figure}

Figure~\ref{fig:ood_cube} reports zero-shot planning success under the OGBench-Cube distribution shifts, complementing the PushT results in the main text
(Figure~\ref{fig:ood_generalization}). SlotContrast-WM and DINO-WM behave similarly in this environment, both staying close to their in-distribution success across all shifts. LeWM is comparatively robust to object-level variations (cube and agent color, cube size) but degrades under scene-level
shifts---background color and camera angle bring it close to the $48\%$ random-policy baseline.

  \begin{table}[h]
      \caption{OGBench-Cube OOD variation values. ``In-dist.''\ is the training-distribution default.}
      \label{tab:variation_values_cube}
      \centering
      \small
      \begin{tabular}{llll}
          \multicolumn{1}{l}{\bf Group} & \multicolumn{1}{l}{\bf Factor} & \multicolumn{1}{l}{\bf In-dist.\ value} & \multicolumn{1}{l}{\bf OOD values}
          \\ \hline \\
          object-level & \texttt{cube.color}         & crimson              & red, blue, yellow \\
                       & \texttt{cube.size}          & $0.020$~m            & $0.012$, $0.028$ \\
                       & \texttt{agent.color}        & purple               & red, black \\
          \\ \hline \\
          scene-level  & \texttt{floor.color}        & slate                & brown, magenta \\
                       & \texttt{camera.angle\_delta}& $(0^\circ, 0^\circ)$ & yaw $\pm 5^\circ$, pitch $\pm 5^\circ$ \\
                       & \texttt{light.intensity}    & $0.6$                & $0.3$, $0.95$ \\
      \end{tabular}
  \end{table}
\section{Extended Related Work}
\label{sec:appendix:related}
\paragraph{Unsupervised object-centric learning.}
Unsupervised object-centric learning~\citep{burgess2019monet, greff2019multi, Locatello2020SlotAttention} aims to decompose a scene into a small set of latent representations, commonly referred to as slots, where each slot captures a distinct object or entity. These representations are typically learned in a self-supervised manner through reconstruction-based objectives. Slot Attention~\citep{Locatello2020SlotAttention} introduced an iterative attention mechanism in which slots compete to explain different regions of an image, enabling effective unsupervised object binding. Subsequent works extended this to video by adding temporal consistency and object dynamics --- SAVi, SAVi++, STEVE~\citep{kipf2022conditional, elsayed2022savi++, Singh2022STEVE}. More recent approaches --- DINOSAUR~\citep{seitzer2023bridging} and the video extensions VideoSAUR, SOLV, and SlotContrast~\citep{zadaianchuk2023objectcentric, aydemir2023self, manasyan2025temporally} --- combine slot-based architectures with pretrained vision foundation models such as DINO to improve robustness and scalability on real-world data.

\paragraph{Object-centric world modeling.}
Early OCWMs paired slot or entity encoders with relational dynamics models, including C-SWM~\citep{kipf2020cswm} and OP3~\citep{veerapaneni2020op3}, predominantly evaluated on synthetic physics. SlotFormer~\citep{wu2023slotformer} replaced graph-net dynamics with an autoregressive Transformer over slots, scaling to richer video datasets. More recent OCWMs target control: FOCUS~\citep{ferraro2025focus} and SOLD~\citep{mosbach2410sold} integrate slots into Dreamer-style model-based RL; Slot-MPC~\citep{spieler2026slot} performs sampling-based MPC in slot space; Dyn-O~\citep{wang2026dyn} adds SAM-supervised slots for Procgen video prediction; OC-STORM~\citep{zhang2025ocstorm} pairs SAM-derived object features with a transformer world model for visually complex domains; and \citet{feng2026learning} factorize representations hierarchically for policy learning.


\end{document}